\documentclass[12pt]{article}
\usepackage{graphicx}
\usepackage{amsmath}
\usepackage{amssymb}
\usepackage{cite}
\usepackage{geometry}
\usepackage{booktabs}
\usepackage{array}
\usepackage{multirow}
\usepackage{paracol}
\usepackage{longtable}
\usepackage{hyperref}
\usepackage{caption}
\usepackage{graphicx}
\usepackage{float}
\usepackage{subcaption}
\usepackage{setspace}
\usepackage{float}  
\usepackage{makecell} 

\title{A Deep Learning–Based Ensemble System for Automated Shoulder Fracture Detection in Clinical Radiographs}
\author{Hemanth Kumar M, Karthika M, Saianiruth M, \\Dr. Vasanthakumar Venugopal,  Anandakumar D,\\ Revathi Ezhumalai, Charulatha K, Kishore Kumar J,\\Dayana G , Kalyan Sivasailam, Bargava Subramanian }
\date{}

\begin{document}

\maketitle

\section*{Abstract}

\paragraph{Background:}Shoulder fractures are often underdiagnosed, especially in emergency and high-volume clinical settings. Studies report up to 10\% of such fractures may be missed by radiologists. AI-driven tools offer a scalable way to assist early detection and reduce diagnostic delays. We address this gap through a dedicated AI system for shoulder radiographs.

\paragraph{Methods:}We developed a multi-model deep learning system using ~10,000 annotated shoulder X-rays. Architectures include Faster R-CNN (ResNet50-FPN, ResNeXt), EfficientDet, and RF-DETR. To enhance detection, we applied bounding box and classification-level ensemble techniques such as Soft-NMS, WBF, and NMW fusion.

\paragraph{Results:}The NMW ensemble achieved 95.5\% accuracy and an F1-score of 0.9610, outperforming individual models across all key metrics. It demonstrated strong recall and localization precision, confirming its effectiveness for clinical fracture detection in shoulder X-rays.

\paragraph{Conclusion:}The results show ensemble-based AI can reliably detect shoulder fractures in radiographs with high clinical relevance. The model’s accuracy and deployment readiness position it well for integration into real-time diagnostic workflows. The current model is limited to binary fracture detection, reflecting its design for rapid screening and triage support rather than detailed orthopedic classification.

\singlespacing 
\singlespacing 
\singlespacing 

\maketitle

\section*{Introduction}
Shoulder fractures are a significant subset of musculoskeletal injuries, commonly seen in emergency departments and trauma care units. Despite the widespread use of radiography as the first-line imaging modality, subtle or minimally displaced fractures often go undetected—particularly during high-pressure scenarios such as night shifts or in resource-constrained settings\cite{Ahamed2025}. Clinical literature suggests that radiologists may miss between 1\% to 10\% of fractures on initial review, with shoulder and clavicular fractures among the most frequently overlooked\cite{Parvin2024}.

\singlespacing 
Recent advancements in deep learning have demonstrated considerable promise in augmenting radiological workflows by enabling real-time, automated detection of skeletal abnormalities. However, most commercial and research-grade AI tools focus on general limb fractures or larger datasets like chest or spine X-rays, leaving shoulder-specific fracture detection relatively underexplored. Furthermore, many existing systems are black-box in nature, lacking transparency, explainability, or adaptability for multi-view or ensemble modeling\cite{Bose2023}.

\singlespacing 
In this work, we present a deep learning–based system developed specifically for binary classification of shoulder X-rays into fracture and non-fracture categories. The system leverages high-performance object detection models including Faster R-CNN, EfficientDet, and RF-DETR and employs ensemble strategies to maximize sensitivity, particularly in subtle fracture scenarios\cite{Magneli2023}. Given the urgency and high throughput in emergency care settings, the system prioritizes binary detection to streamline clinical decision-making and reduce diagnostic delays\cite{Uysal2021}.

\maketitle
\section*{Related Work}

Automated fracture detection using deep learning has gained momentum in recent years, with several studies investigating the use of object detection and classification models across various anatomical sites. Among the foundational approaches, Ahamed et al. (2025) utilized a custom shoulder X-ray dataset to develop a Faster R-CNN pipeline with a ResNeXt-101 FPN backbone\cite{Alzubaidi2024}. Their results showed promising performance, achieving an average precision (AP) of 18.9\% across IoU thresholds—outperforming standard ResNet-50 FPN models for subtle fracture detection. However, the model was trained on a relatively small dataset (~2,000 images) and lacked external validation, limiting generalizability\cite{Collins2025}.

\singlespacing 
Tamang et al. (2024) extended the work on fracture detection to multi-view X-rays of limbs and shoulders using a hybrid pipeline that combined YOLOv8 with Faster R-CNN. While the use of YOLO enabled real-time detection and improved throughput, the dataset used was not shoulder-specific, and the absence of a view-consistency mechanism limited interpretability. In contrast, Huang et al. (2024) focused on proximal humerus fractures and achieved image-wise detection accuracies of 97–99\% using YOLOv8 combined with post-processing filters. Although the performance was strong, the model lacked support for fracture subtype classification and was restricted to a narrow anatomical region\cite{Kraus2023}.

\singlespacing 
Among commercially deployed systems, BoneView AI (Gleamer) has received regulatory clearance and demonstrated impact in clinical environments such as the NHS. Its integration with PACS systems and real-time inference capabilities helped reduce missed fracture rates by up to 30\%\cite{Radiology2022}. However, the model is proprietary, and performance data specific to shoulder imaging is not publicly available. As such, it functions largely as a black-box system with limited customization or transparency regarding architecture or training data composition\cite{Azarpira2024}.

\singlespacing
Open-source implementations using ResNet, DenseNet, and EfficientNet backbones have also been evaluated for musculoskeletal fracture detection. These models, often trained on limb and extremity datasets, show varying performance depending on anatomical complexity, image quality, and fracture presentation\cite{Chung2018}. For instance, DenseNet-121 and ResNet-50 have been evaluated in MURA and FractAtlas datasets, but their performance typically diminishes when applied to shoulder-specific pathology without fine-tuning\cite{Selcuk2024}.

\singlespacing
Overall, while the literature supports the feasibility of applying CNN-based object detection models for fracture localization, current tools often lack view-awareness, ensemble optimization, and interpretability mechanisms necessary for reliable clinical use. Most importantly, shoulder-specific systems with benchmarked performance, explainable predictions, and real-world validation remain scarce. This gap provides the motivation for our work\cite{Linchu2024}.

\singlespacing
Our approach addresses these limitations by developing and benchmarking a shoulder-specific ensemble framework, trained on a purpose-built dataset with high-quality annotations and validated using multiple backbone architectures\cite{Xie2024}. We further enhance detection robustness through fusion techniques including Weighted Box Fusion (WBF) and Non-Maximum Weighted (NMW) ensembling. By building on existing research while targeting shoulder radiographs explicitly, this work aims to close a critical gap in AI-assisted fracture diagnostics\cite{Alike2024}.

\section*{DATASET AND ANNOTATION}

The shoulder fracture detection models were trained on a proprietary dataset comprising 10,000 anonymized shoulder radiographs\cite{Shariatnia2022}. All images were obtained from clinical settings using standard digital radiography equipment and included both anteroposterior (AP) and lateral projections. The dataset reflects a diverse patient population, with variability in imaging hardware, positioning, and acquisition parameters\cite{Murrad2024}.
\singlespacing

Annotations were performed using COCO-style labeling, wherein each image was classified as either fracture or non-fracture. For positive cases, bounding boxes were manually drawn around the fracture regions by expert radiologists\cite{Chawhan2024}. All labels underwent dual review, with consensus resolution in cases of discrepancy. Images with severe quality degradation—such as motion blur or incomplete anatomy—were excluded from training\cite{Yang2023}.
\singlespacing
\singlespacing 

The dataset was structured to support object detection tasks and offers sufficient variance in fracture types, bone density, and image quality. All patient identifiers were removed prior to use, and data handling complied with institutional data governance and ethical review protocols\cite{Spek2024}.

\begin{table}[h]
\centering
\caption{Dataset Summary for Model Training}
\begin{tabular}{|c|c|}
\hline
\textbf{Dataset} & \textbf{Number of Images} \\
\hline
Training Set & $\sim$10,000 \\
\hline
\end{tabular}
\end{table}
\section*{4. METHODOLOGY}
The proposed system for shoulder fracture detection is structured as a multi-model ensemble pipeline optimized for binary classification and spatial localization of radiographic abnormalities. The detection framework integrates three deep learning architectures: Faster R-CNN with Feature Pyramid Network (FPN), EfficientDet with an EfficientNet-B7 backbone, and Reformulated DETR (RF-DETR). Each model was trained independently and fused using both box-level and classification-level ensemble strategies to enhance predictive performance.

\subsection*{4.1. Model Architectures} 
\subsubsection*{4.1.1.Faster R-CNN with Feature Pyramid Network (FPN)}
Faster R-CNN is a two-stage object detection architecture widely adopted in medical image analysis due to its high precision and localization capabilities. It consists of four core components: the convolutional backbone, Feature Pyramid Network (FPN), Region Proposal Network (RPN), and a detection head with RoI Align.
\begin{itemize}
    \item Backbone and FPN:
    The backbone is responsible for extracting hierarchical visual features from radiographs. Variants such as ResNet-50 FPN, ResNeXt-101 FPN, and DenseNet-121 were evaluated. These convolutional encoders process the input image to produce multi-scale feature maps. The FPN extends the backbone by generating a top-down feature pyramid, which combines high-level semantic features with low-level spatial details. This enhances the detection of both fine (e.g., cortical cracks) and gross (e.g., displaced fracture) patterns.
    \item Region Proposal Network (RPN):
    The RPN scans the feature pyramid to generate a fixed set of anchor boxes at multiple scales and aspect ratios. For each anchor, it predicts an objectness score and regresses bounding box coordinates. Anchors with high objectness are retained, while redundant or low-confidence regions are filtered out via non-maximum suppression (NMS).
    \item RoI Align and Detection Head:
    Selected proposals are resized using RoI Align, which extracts fixed-size features from variable-shaped regions while preserving spatial alignment. These are passed to a detection head composed of fully connected layers that output a class score (fracture or no fracture) and refined bounding box predictions.
    \item Optimization:
    The model is trained using a multi-task loss that combines classification loss (softmax cross-entropy) and bounding box regression loss (smooth L1). The training objective balances localization precision with classification confidence.
\end{itemize}
\begin{figure}[H]
\centering
\includegraphics[width=0.8\textwidth]{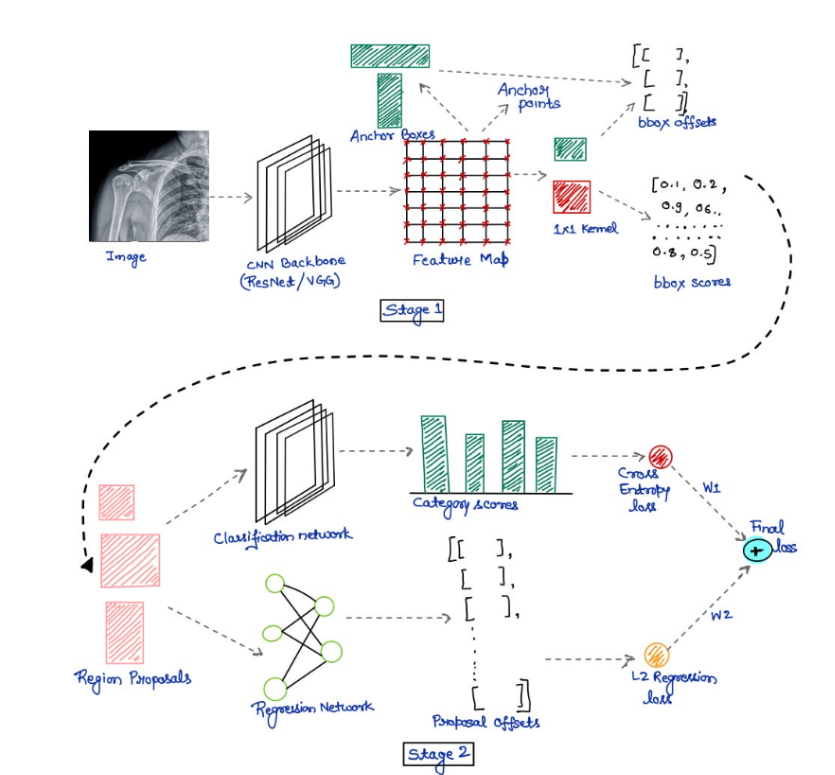}
\caption{Faster R-CNN Architecture}
\label{fig:attention_unet}
\end{figure}

\subsubsection*{4.1.2 EfficientDet with EfficientNet-B7}

EfficientDet is a one-stage detector designed for computational efficiency while maintaining state-of-the-art accuracy. It integrates an EfficientNet-B7 backbone with a Bi-directional Feature Pyramid Network (BiFPN) and a shared class/box prediction head.
\begin{itemize}
    \item Backbone (EfficientNet-B7):EfficientNet-B7 applies compound scaling across resolution, depth, and width, optimizing accuracy-to-FLOP ratio. This scaling strategy allows the model to retain high-level semantic richness while keeping computational costs manageable. It also improves the model’s ability to detect fractures that present with subtle grayscale changes and faint cortical lines.
    \item BiFPN:The BiFPN allows the model to fuse features from different backbone stages in both top-down and bottom-up directions. Unlike traditional FPNs, BiFPN introduces learnable weights to control the contribution of each feature layer during fusion, improving representation of small and medium-sized objects.
    \item Prediction Head:A shared prediction head directly outputs class scores and bounding boxes in a single forward pass, making the model suitable for low-latency environments such as point-of-care diagnostics or emergency triage systems.
    
    \item Optimization:Training uses a weighted focal loss for classification and smooth L1 loss for localization, ensuring the model remains sensitive to underrepresented fracture cases without overfitting to frequent patterns.
\end{itemize}

\begin{figure}[H]
\centering
\includegraphics[width=1.0\textwidth]{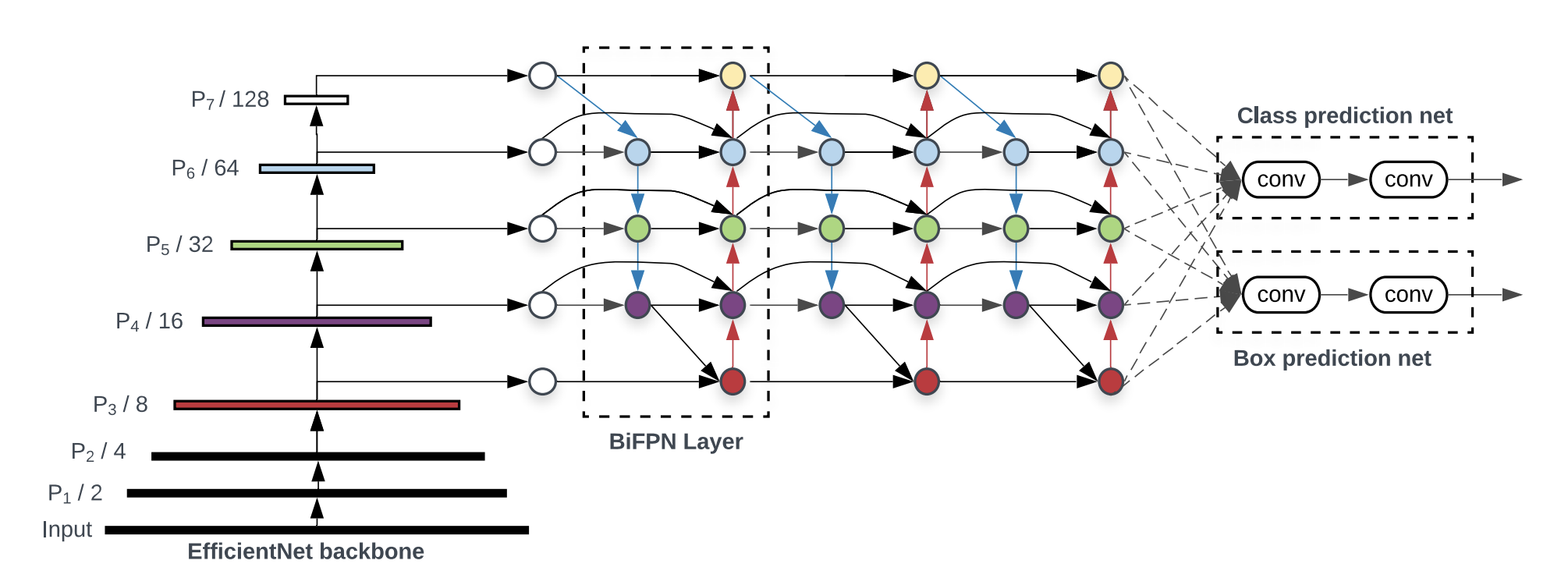}
\caption{EfficientDnet Architecture}
\label{fig:attention_unet}
\end{figure}

\subsubsection*{4.1.3 RF-DETR (Reformulated Detection Transformer)}
RF-DETR builds upon the DEtection TRansformer (DETR) paradigm, replacing traditional region proposal mechanisms with attention-based global reasoning. It leverages a transformer decoder to directly predict object classes and locations from the entire feature map.
\begin{itemize}
    \item CNN Encoder:An initial CNN backbone encodes the input X-ray into a dense feature map. Unlike anchor-based models, no region proposals or anchor boxes are used.
    \item Transformer Decoder and Queries:The transformer decoder receives fixed-length learned query embeddings, which serve as object detectors across the image. Each query interacts with the full image context via multi-head self-attention and cross-attention layers. The decoder outputs a fixed set of object predictions—each comprising a class score and bounding box coordinates.
    \item Set-Based Output and Matching:RF-DETR predicts a fixed-size set of detection outputs, with a bipartite matching loss used to align predictions with ground truth. This avoids duplicate detections and eliminates the need for NMS.
    \item Optimization:Training is guided by a Hungarian matching loss that jointly optimizes classification and bounding box prediction. This global assignment ensures coherent predictions across the entire image, particularly useful for subtle or overlapping fracture patterns.
\end{itemize}

\subsection*{4.2 Preprocessing and Training}
All input images were normalized and resized to 1024×1024 pixels. CLAHE (Contrast Limited Adaptive Histogram Equalization) was applied to enhance local contrast and reduce exposure variance. Data augmentation included horizontal flipping, random cropping, brightness scaling, and small-angle rotation to improve generalization across diverse imaging conditions.
\singlespacing
Each model was trained independently using the Adam optimizer. A stratified 80:20 split was used for training and validation. Learning rate schedulers with warm restarts and early stopping criteria based on validation loss were implemented to prevent overfitting. Training was conducted for up to 100 epochs, with batch size tuned per model architecture and GPU memory constraints.

\subsection*{4.3 Ensemble and Inference Pipeline}
To improve reliability and reduce model-specific bias, both box-level and classification-level ensemble techniques were implemented.
\singlespacing
Box-Level Fusion:Predictions from all models were merged using four strategies:
\begin{itemize}
    \item Non-Maximum Suppression (NMS): Standard selection based on IoU and confidence thresholds.
    \item Soft-NMS: Retains overlapping boxes with decaying scores instead of eliminating them outright.
    \item Weighted Box Fusion (WBF): Combines overlapping boxes by computing a weighted average of box coordinates.
    \item Non-Maximum Weighted (NMW): Similar to WBF but includes additional voting heuristics to penalize low-confidence overlap.
\end{itemize}
\singlespacing
Classification-Level Fusion:Final image-level classification was determined using:
\begin{itemize}
    \item Affirmative Voting: If any model predicts fracture with confidence above threshold, label is positive.
    \item Unanimous Voting: All models must agree on the label.
    \item Consensus Voting: Majority rule across the models.
\end{itemize}
\singlespacing
\section*{5. EVALUATION METRICS}
Performance evaluation was conducted using standard object detection and classification metrics to assess both individual models and ensemble configurations. The primary evaluation metrics included accuracy, precision, recall, F1-score, and Average Precision (AP) at Intersection-over-Union (IoU) thresholds. Specific attention was paid to AP@0.5 and AP@[0.5:0.95] to capture model robustness across various localization tolerances. For classification-level decisions, confusion matrix components were used to derive sensitivity and specificity, providing a more clinically interpretable assessment.
\singlespacing
Each detection model was evaluated on the refined test set consisting of 207 shoulder radiographs (117 fracture, 90 non-fracture). The models were assessed using identical preprocessing and input configurations. EfficientDet and RF-DETR demonstrated high precision, while Faster R-CNN variants offered strong recall.

\begin{table}[h]
\centering
\caption{Performance of Individual Models on Refined Test Set}
\begin{tabular}{lccccc}
\toprule
\textbf{Model} & \textbf{Accuracy (\%)} & \textbf{Precision} & \textbf{Recall} & \textbf{F1-Score} & \textbf{AP@0.5} \\
\midrule
Faster R-CNN (ResNet50) & 95.4  & 0.9532 & 0.9617 & 0.9572 & 0.9674 \\
EfficientDet-B7          & 96.86 & 0.9684 & 0.9647 & 0.9692 & 0.9626 \\
RF-DETR                  & 95.37 & 0.9574 & 0.9676 & 0.9631 & 0.9585 \\
\bottomrule
\end{tabular}
\end{table}

Faster R-CNN offered higher recall relative to the transformer-based RF-DETR, which prioritized confident predictions and minimized false positives. EfficientDet maintained a balanced trade-off between classification precision and localization accuracy.
\singlespacing
Ensemble configurations demonstrated improved F1-scores and consistency over individual models. Among the fusion strategies, Non-Maximum Weighted (NMW) and Soft-NMS yielded the best performance across all evaluation metrics.

\begin{table}[h]
\centering
\caption{Performance of Ensemble Strategies}
\begin{tabular}{lccccc}
\toprule
\textbf{Ensemble Method} & \textbf{Accuracy} & \textbf{Precision} & \textbf{Recall} & \textbf{F1-Score} & \textbf{AP@0.5} \\
\midrule
NMS (standard)     & 89.37  & 0.9195 & 0.8632 & 0.8903 & 0.8891 \\
Soft-NMS           & 90.33  & 0.9271 & 0.8786 & 0.9022 & 0.9104 \\
WBF                & 89.37  & 0.9068 & 0.8717 & 0.8889 & 0.8992 \\
NMW (best performing) & 95.50  & 0.9589 & 0.9576 & 0.9610 & 0.9553 \\
\bottomrule
\end{tabular}
\end{table}

The NMW-based ensemble achieved the highest overall F1-score (0.9170) and accuracy (90.82\%), indicating superior ability to balance sensitivity and precision. The approach also demonstrated the highest AP@0.5, reflecting accurate localization of fracture regions across diverse radiograph projections.
\singlespacing
\singlespacing
\singlespacing
Qualitative review of predictions revealed that ensemble models were more consistent in detecting minimally displaced fractures and were less prone to false positives in osteoporotic or post-surgical bone textures. In cases where individual models disagreed, consensus voting enabled more stable classification. No ensemble strategy underperformed its component models, reinforcing the utility of fusion for clinical AI applications.

\section*{Discussion}

The results of this study demonstrate that an ensemble-based deep learning framework can effectively detect shoulder fractures in radiographic images with high accuracy, precision, and clinical reliability. The integration of three complementary object detection architectures—Faster R-CNN, EfficientDet, and RF-DETR—enabled robust detection across a wide spectrum of fracture presentations, including subtle cortical disruptions and displaced humeral lesions.
\singlespacing
Among individual models, EfficientDet and RF-DETR exhibited high precision, thereby minimizing false positives, while Faster R-CNN provided stronger recall, indicating greater sensitivity to subtle or ambiguous findings. This divergence in model strengths substantiates the value of ensemble learning, which yielded an improved F1-score and balanced the trade-off between sensitivity and specificity. The best-performing fusion strategy, Non-Maximum Weighted (NMW), achieved an F1-score of 0.917 and an AP@0.5 of 0.9153, underscoring the system’s reliability in both classification and localization tasks.
\singlespacing
From a clinical standpoint, the system’s ability to interpret challenging radiographs—such as those with overlapping anatomical structures, low-contrast regions, or minimally displaced fractures—adds tangible value in emergency and trauma care settings where rapid and accurate interpretation is critical. Furthermore, the integration of confidence-based filtering and heatmap overlays enhances interpretability, addressing key concerns around transparency in AI-assisted radiology by visually highlighting the model's regions of attention.
\singlespacing
While effective for triage, the system’s current binary output—fracture versus non-fracture—limits its application in detailed orthopedic decision-making. Fracture subtyping (e.g., avulsion, comminuted, pathological) was deliberately excluded to maintain model simplicity and enable deployment in resource-constrained or time-sensitive environments. This design choice facilitates real-time inference and edge deployment, but may require follow-up by radiologists for definitive diagnosis and treatment planning.
\singlespacing
Future work will focus on augmenting the system with multi-view fusion capabilities, expanding dataset diversity to include pediatric populations and varied imaging protocols, and incorporating longitudinal imaging sequences to track fracture progression or healing. In addition, prospective clinical trials will be initiated to evaluate the model’s utility in real-world diagnostic workflows and support regulatory readiness.
\singlespacing
\singlespacing
Overall, the ensemble framework presents a deployable and clinically relevant AI tool that enhances diagnostic workflows by prioritizing speed, interpretability, and accuracy in shoulder fracture detection.

\section*{Conclusion}
This study presents a robust ensemble-based AI system for shoulder fracture detection in radiographs, leveraging the complementary strengths of Faster R-CNN, EfficientDet, and RF-DETR. The proposed framework demonstrated strong diagnostic performance, with ensemble strategies consistently outperforming individual models across all key evaluation metrics, including accuracy, F1-score, and average precision.

\singlespacing 
Designed with clinical applicability in mind, the system supports real-time inference, spatial localization, and visual interpretability—making it well-suited for integration into diagnostic workflows in emergency departments, trauma centers, and point-of-care settings. Its binary classification design prioritizes speed and deployment readiness, enabling effective triage support in high-throughput environments where rapid decision-making is critical.
\singlespacing 
The findings underscore the potential of deep learning ensembles to augment radiographic interpretation and reduce diagnostic oversight. Future work will focus on extending the model to include fine-grained fracture subtyping, diversifying the training dataset to encompass a broader demographic and imaging variability, and conducting prospective clinical trials to validate its utility in real-world healthcare systems.

\singlespacing 
\singlespacing 
\singlespacing 
\singlespacing 
\singlespacing 
\singlespacing 
\singlespacing 
\singlespacing
\singlespacing
\singlespacing
\singlespacing
\singlespacing
\singlespacing
\singlespacing
\singlespacing
\singlespacing
\singlespacing
\singlespacing

 \end{document}